\documentclass[letterpaper, 10 pt, journal, twoside]{ieeetran}
\usepackage{geometry}
\IEEEoverridecommandlockouts   

\usepackage{graphics} 
\usepackage{epsfig} 
\usepackage{amsmath} 
\usepackage{amssymb}  
\usepackage{multirow}
\usepackage{url}
\usepackage{siunitx}
\usepackage{color} 

\usepackage[numbers]{natbib}
\usepackage[pageanchor=true,plainpages=false, pdfpagelabels, bookmarks,bookmarksnumbered,hidelinks]{hyperref}
\hypersetup{nolinks=true}

\newcommand{\alex}[1]{\textcolor{black}{#1}}
\newcommand{\alexnew}[1]{\textcolor{black}{#1}}

\usepackage{titlesec}

\title{Femtosecond laser fabricated nitinol living hinges for millimeter-sized robots}

\author{Alexander Hedrick$^{1}$, Heiko Kabutz$^{1}$, Lawrence Smith$^{1}$, Robert MacCurdy$^{1}$, and Kaushik Jayaram$^{1,*}$
\thanks{Manuscript received: October, 24, 2023; Revised February, 4, 2024; Accepted March, 28, 2024.}
\thanks{This paper was recommended for publication by Editor Xinyu Liu upon evaluation of the Associate Editor and Reviewers' comments. This work is partially funded through grants from the Paul M. Rady Mechanical Engineering Department.}
\thanks{$^{1}$Paul M. Rady Department of Mechanical Engineering, University of Colorado Boulder} 
\thanks{$^{*}${For correspondence, \tt\footnotesize kaushik.jayaram@colorado.edu}}%
\thanks{Digital Object Identifier (DOI): see top of this page.}
} 

\begin{document}
\maketitle

\begin{abstract}

Nitinol is a smart material that can be used as an actuator, a sensor, or a structural element, and has the potential to significantly enhance the capabilities of microrobots. Femtosecond laser technology can be used to process nitinol while avoiding heat-affected zones (HAZ), thus retaining superelastic properties.  In this work, we manufacture living hinges of arbitrary cross sections from nitinol using a femtosecond laser micromachining process. We first determined the laser cutting parameters, {\boldmath\SI{4.1}{Jcm^{-2}}} fluence with {\boldmath\SI{5}{}} passes for {\boldmath\SI{5}{\micro \meter}} ablation, by varying laser power level and number of passes. Next, we modeled the hinges using an analytical model as well as creating an Abaqus finite element method, and showed the accuracy of the models by comparing them to the torque produced by eight different hinges, four with a rectangular cross section and four with an \alex{elliptic notch} cross section. Finally, we manufactured a prototype miniature device to illustrate the usefulness of these nitinol hinges: a piezoelectric actuated robotic wing mechanism, and we characterized its performance.

\end{abstract}


\IEEEpeerreviewmaketitle

\section{Introduction}
\label{sec:intro}

\IEEEPARstart{M}{iniature} robots are becoming increasingly relevant for a variety of real-world applications including search-and-rescue \cite{jayaram2016cockroaches}, high-value asset inspection and maintenance \cite{de2018inverted}, environmental monitoring \cite{hu2018small} and healthcare \cite{nelson2010microrobots}. 
While tremendous progress has been made in the rapid prototyping of millimeter-scale devices, especially with respect to 3D printing \cite{wallin20183d} and laminate stack manufacturing \cite{cho2009review}, both have significant issues currently limiting wider adoption. 3D printing is limited by choice of materials and scale of printing, trading off resolution of fine features with overall build volume \cite{wallin20183d}. 
On the other hand, laminate stack manufacturing allows combining multiple materials, but the typical design process is often complex and is limited to the planar processing of thin films \cite{wood2008microrobot}. 
For example, to make a symmetrically bending hinge, one typically needs to combine at least five layers - two structural, two adhesive, and one flexible \cite{sreetharan2012monolithic}. Moreover, the flexible layer may contain through-hole features but cannot support 3D contouring. This creates bidirectional flexures which are ideal for bending about a single axis but have off-axis compliance that cannot be tuned easily, resulting in less-than-ideal robot dynamics. 
For example, flexures of climbing \cite{de2018inverted}, high speed running \cite{doshi2019effective}, and flying robots have a small elastic deflection range (0.8\% elastic strain limit for polyimide Kapton \cite{yu_kapton_2004}) and limited lifetime (100s for the flying robot from Kim, et al. \cite{kim_laser-assisted_2023}) resulting in poor overall performance. 
Therefore, addressing the grand challenge of developing novel materials and new fabrication schemes \cite{yang2018grand} which increase the available choices to robotic designers is instrumental in realizing the next generation of highly capable robots \cite{zhang_advanced_2023}.

\begin{figure}[tp!]
    \centering
    \includegraphics[width = 0.4\textwidth]{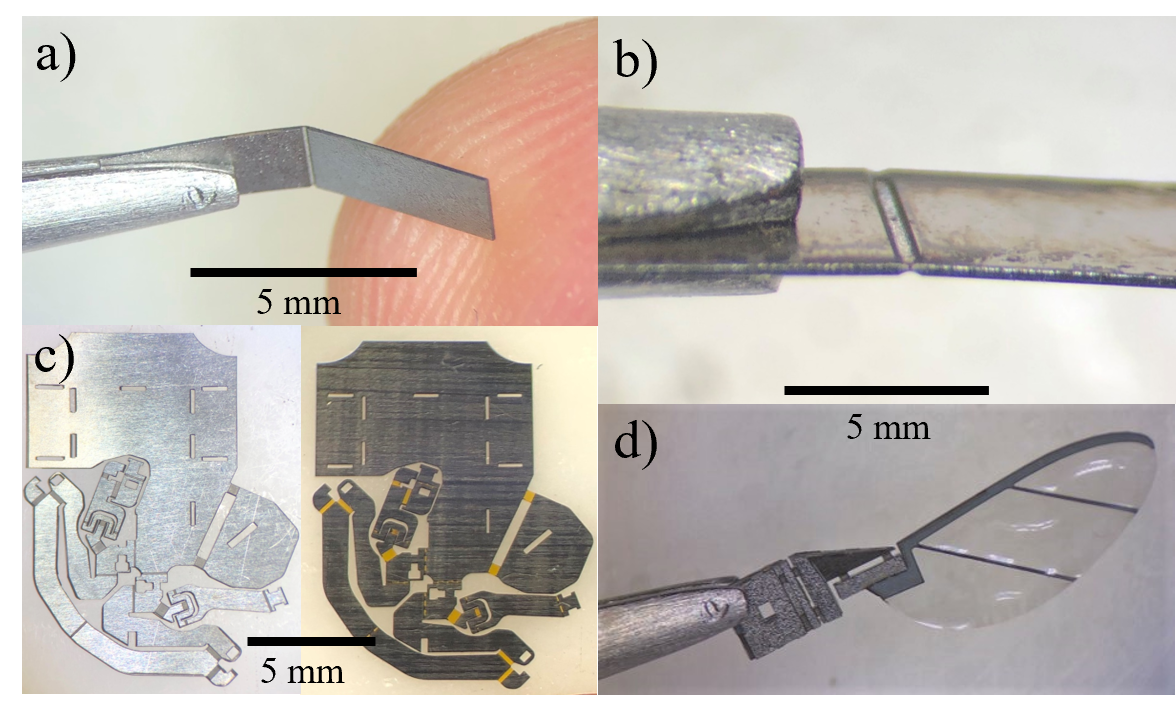}
    \caption{(a): One of four rectangular cross section hinges used to characterize hinge torque and validate hinge models. (b): Living hinge manufactured with cutouts on both sides of bulk material, as used in the wing mechanism. (c): 5 bar mechanism for a millirobot manufactured from nitinol (left) and from the traditional carbon fiber and Kapton (right) \cite{kabutz2023design}. (d): Wing mechanism manufactured from carbon fiber, nitinol, and mylar.}
    \label{fig:hinge_main}
\end{figure}

Our group \cite{kabutz2023design, kabutz2023mclari} is specifically interested in exploring the potential of using nitinol for microrobotics \cite{palagi2018bioinspired,soto2021smart}.
Nitinol, a nickel-titanium alloy, has a wide variety of unique properties that make it useful for many different applications at larger scales. Some of these properties include superelasticity (4-10\% elastic strain limit), shape memory, high kink and fatigue resistance, non-linear stiffness, and biocompatibility. These properties, especially biocompatibility, have led to nitinol becoming increasingly used in medical devices, such as for stents, filters, scissors, grabbers, and more \cite{stockel_nitinol_nodate,  duerig_overview_1999}. Additionally, nitinol can be used as an actuator (shape memory) or high-deformation flexure (superelasticity) in various applications in the robotics \cite{yang_88-milligram_2020}, precision medicine \cite{york_microrobotic_2021}, aerospace \cite{chaudhari_review_2021}, and automotive \cite{mertmann_nonmedical_2004} industries. 
We believe a custom-machined three-dimensional nitinol flexure fabricated at millimeter sizes can potentially address many of the concerns described previously to enable further miniaturization of millimeter-scale devices beyond their current sizes \cite{jayaram2020scaling}. 
So far, nitinol has been demonstrated to be effectively used as a structural material \cite{liu2013compliant, haberland2013visions}, an actuator \cite{mertmann_nonmedical_2004, costanza2010nitinol} and/or a sensor \cite{srivastava2016design, ruth2021design} in miniature designs. Such multifunctionality makes it particularly attractive  for monolithic fabrication and miniaturization of microscale devices.


However, nitinol is a notoriously difficult material to machine. Rapid work hardening can grind down mechanical tools or require a manufacturer to anneal the material, a time-consuming process \cite{hodgson_nitinol_2000}. Applying heat shaping techniques to process nitinol is a possibility, but often expensive and complex \cite{hodgson_nitinol_2000}. 
\alex{Furthermore, nitinol has a relatively small critical crack size, so fatigue life is driven by crack initiation instead of crack propagation. Thus, increased care must be taken to reduce the likelihood of crack formation while working with nitinol \cite{mahtabi_fatigue_2015}.}

Considering the above challenges, current processing methods include abrasive jet machining (AJM), which is an effective manufacturing technique for this material as recently demonstrated, but it cannot yet accurately produce the high-resolution (e.g. few microns) complex features (e.g. geometries with varying cross sections) essential for microrobotic devices \cite{york_nitinol_2019}. \alex{Focused ion beam (FIB) milling has been used successfully to manufacture nitinol actuators from nitinol wire, but FIB cannot produce complex 2.5D features easily \cite{lee_shape_2018}. Similarly, photo-chemical etching is a cheap and accurate way to manufacture nitinol devices \cite{shayan_overview_2015}. However, limitations of this process include the inability to create complex 2.5D or 3D cutouts and long processing times relative to laser machining.
Alternative approaches for creating complex contours for microrobotic applications \cite{de2018inverted, jayaram2020scaling, kabutz2023design, kabutz2023mclari} include laser photomicromachining.
In these processes, thermal diffusion is proportional to the square root of pulse width \cite{zheng_femtosecond_nodate} and utilizing micro- or nano- second pulses, a prevalent option for typical laser cutters, creates heat-affected zones (HAZ). HAZ negatively affect the superelasticity and/or shape memory of the bulk material thus making the material unusable for high-performance applications \cite{mwangi_nitinol_2019}.
However, with progress in laser technology, the use of femtosecond pulses has recently emerged as a rapid manufacturing technique to process nitinol accurately while minimizing HAZ \cite{mwangi_nitinol_2019, uppal_micromachining_2008, huang_femtosecond_2004, pfeifenberger2017use}. 
Due to the incredibly short laser pulse width ($\sim 100$ fs), evaporation is dominant and heat conduction can be effectively ignored \cite{huang_femtosecond_2004}. This process, called 'cold ablation', evaporates the material rather than melting it away resulting in machined material retaining its bulk properties \cite{pfeifenberger2017use}.} Despite the benefits of machining nitinol with a femtosecond laser, few examples of this manufacturing method exist in the literature \cite{uppal_micromachining_2008, pfeifenberger2017use}, primarily due to the expensive cost of femtosecond lasers and associated optics. However, with this technology becoming significantly cheaper in the last five years, femtosecond laser micromachines are starting to become commercially available (e.g. 6D Laser, LLC\footnote{\url{www.6dlasers.com}}). 

In this work, we explore the idea of creating nitinol living hinges using a custom femtosecond laser micromachine. This motivation is driven by the knowledge that living hinges are one of the fundamental building blocks of compliant mechanisms and in general reduce the size, part count, and energy losses compared to traditional mechanisms \cite{hopkins2007design, shaw2019computationally, mccarthy2023design}. 
We demonstrate laser manufacturing of nitinol living hinges with arbitrary cross sections. We then characterize these hinges by measuring their torque-angle relationship. 
We model the fabricated living hinges using theory and finite element analysis and show that our experimental data match these models. 
Finally, we design and test a miniature robotic wing mechanism as a demonstration of our process and highlight its potential in the rapid prototyping of insect-scale flying robots. 
We conclude by presenting future development, as we expect to further improve the manufacturing process for the nitinol hinges and integrate them into millimeter-scale robot and medical devices that would benefit from hinges with high displacement.

\section{Femtosecond Laser Micromachining}
\label{sec:manu}

In this section, we detail the fabrication process used to manufacture nitinol living hinges. We fabricated our hinges using a \alex{\SI{100}{\micro m} thick superelastic nitinol sheet with a transition temperature of \SI{-10}{\degree C} (Nexmetal Corp.). All experiments were performed at room temperature.}

\subsection{Laser Micromachining Setup}
Our custom laser micromachine (6D Laser, Fig. \ref{fig:workflow}a), features a Light Conversion CARBIDE-CB5 Femtosecond laser cutter with UV harmonics. This setup has an ALIO 6-D Hybrid Hexapod stage with sub-micron precision and an integrated SCANLAB excelliSCAN galvo positioning system. With standard optical settings, our spot size is \SI{8}{\micro m} and our field of view is \SI{45}{\milli m} $\times$ \SI{45}{\milli m}, but can be easily tuned (i.e. trading off smaller spot for smaller field of view). To machine samples, we mount them onto an adhesive (Gel-Pak x8 WF Film) to ensure a flat surface while laser cutting. 
Custom software (Direct Machining Control) reads our drawings (as DXFs or STLs) and generates control signals to drive the galvanometer and stage height for laser rastering or marking as desired.  

\subsection{Processing Parameters}
\begin{figure*}[htb!]
    \centering
    \includegraphics[width=0.9\textwidth]{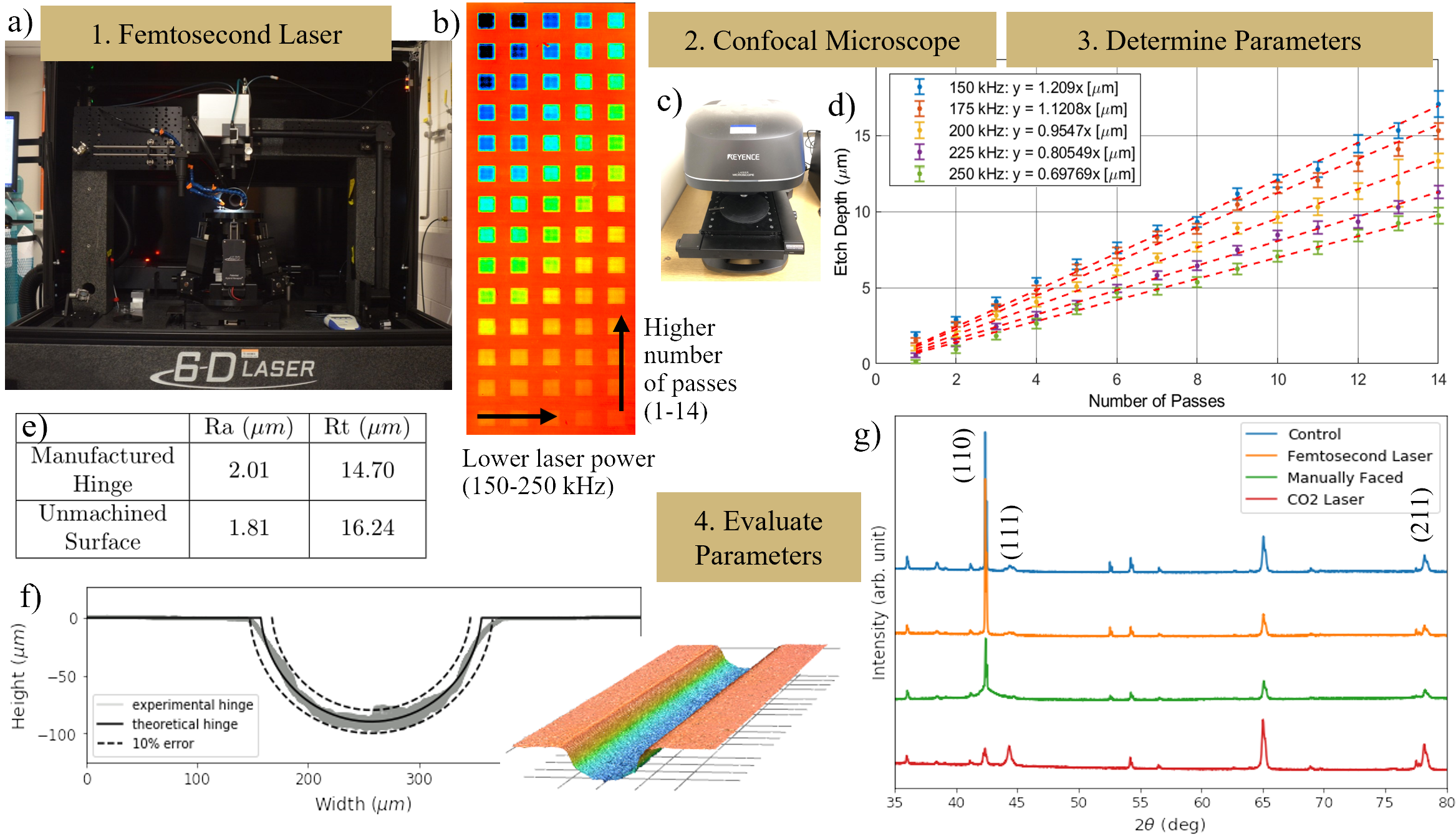}
    \caption{Workflow for establishing laser cutting parameters. (a): Femtosecond laser micromachine. (b): Array of squares used for characterizing the nitinol sheet cut by our laser. The horizontal axis of the array includes the laser power levels 150-250 kHz in increments of 25 kHz. The vertical axis runs from 1-14 passes. Each square has a side length of 200 $\mu m$ and is filled with a grid of cuts spaced 5 $\mu m$ apart. (c): Confocal microscope. (d): Etch depth as a function of the number of passes for each of the five power levels tested along with lines of best fit. (e): Roughness comparison between the manufactured hinge and unmachined surface. (f): Plot comparing the 2 mm wide model hinge cross section with a manufactured one to demonstrate its accuracy and potential for realizing miniature features. (g) Graph of Intensity as a function of 2$\theta$ from XRD test for four different samples with major austenite and martensite peaks labelled. (110) and (211) peaks correspond to austenite and (111) corresponds to martensite.}
    \label{fig:workflow}
\end{figure*}

To create living hinges, we rastered away excess material from the nitinol base substrate. To obtain accurate processing parameters, we build on the procedure detailed in our earlier work \cite{hayes2022rapid, hayes2024scaling}. To summarize, we first ran a custom recipe to create an array of squares with varying laser power levels (repetition frequency) and a number of passes. After ablation, we characterized the sample in a confocal microscope (Keyence VK-X 3D Surface Profiler) set to 0.5 $\mu m$ tolerance and measured the etch depth as a function of repetition frequency and passes. Fig. \ref{fig:workflow}b shows a sample output from the profiler, and Fig. \ref{fig:workflow}c shows the profiler itself.


We compiled the characterization data from the above tests as a scatter plot with lines of best fit in Fig. \ref{fig:workflow}d. 
Based on these data, we determined our "strong ablation range" \cite{uppal_micromachining_2008} to be 200 kHz (equivalent to a fluence of \SI{4.1}{Jcm^{-2}}) and 5 passes resulting in 5 $\mu m$ thick ablations per layer. \alex{We chose these settings based on a trade-off between material quality and manufacturing time. }
We used the above combination of parameters for the rest of the manuscript unless mentioned otherwise. 


\subsection{Evaluation of Laser Processing}

\alex{To confirm our that the our chosen femtosecond laser processing parameters produces negligible HAZ, we performed an X-Ray diffraction (XRD) material characterization test to observe the crystallographic structure of nitinol. Our test contained four samples: a non-machined control, a femtosecond processed sample using our parameters obtained above, a sample processed using a CO2 laser cutter with power on the scale of 10 W, and a sample that was faced with a manual mill. The symmetric $\theta/2\theta$ XRD scans ranged from 30-100 degrees and used a Copper K-$\alpha$ source. A 5 mm wide beam was used for the control and a 2 mm beam for the other three test samples. Fig. \ref{fig:workflow}g shows the XRD results for the four samples with the major austenite and martensite peaks labelled according to their Miller indices \cite{ou_manufacturing_2018}. 
The control and femtosecond laser processed samples have very similar intensities for both austenite and martensite, suggesting that the femtosecond process has negligible effects on the crystal structure of the nitinol. However, the manually faced and CO2 laser processed samples have much lower intensities for austenite and, for the CO2 laser sample, higher intensities for martensite, indicating that these processes do change the underlying crystal structure and are therefore non-ideal manufacturing techniques which supports observations from prior studies \cite{hodgson_nitinol_2000, mwangi_nitinol_2019}.}
\alex{Additionally, to quantify the quality of our manufacturing process, we fabricated an elliptic notch living hinge (depth of 90 $\mu m$) with the ideal parameters as determined above using the confocal microscope described previously. Fig. \ref{fig:workflow}f plots the expected cross section with the manufactured one over the hinge width of 2 mm. This figure shows our ability to create hinges with a tolerance below 10\% of the depth of the cut, and, excluding a few outliers, the large majority of the hinge is within $\pm 5 \mu$m of the designed cut.}

\alex{To evaluate the surface finish of the laser processing technique which is noted to have a significant effect on fatigue performance \cite{zheng_femtosecond_nodate}, we measured  the roughness of the manufactured hinge and compared it to the original roughness of the material. We chose five random cross sections from a confocal image of a 2 mm long hinge and five from the unmachined surface of the nitinol. From these cross sections, we computed the mean roughness (Ra) and total roughness (Rt) and record their averages in Fig. \ref{fig:workflow}e.} 

\alex{These measured Ra values are comparable to those reported in previous studies \cite{zheng_femtosecond_nodate}. Furthermore, the difference in average roughness of the machined vs. unmachined surface is determined to be 0.2 $\mu m$, indicating that the laser cutting process did not significantly increase the roughness of the hinge surface and thus does not have a detrimental effect on performance.} 

\alex{In Section \ref{sec:disc}, we discuss methods for improving the quality (reducing tolerance) and surface finish (reducing roughness) even further using additional processing techniques.}

    


\section{Hinge Design and Characterization}
\label{sec:test}


We decided to manufacture and test \alex{two different types of living hinges} with varying thicknesses: rectangular cross section, shown in Fig. \ref{fig:analysis}a and \alex{elliptic notch}, shown in Fig. \ref{fig:analysis}b. The table in Fig. \ref{fig:analysis}c shows the dimensions and material properties of the hinges. Ease of fabrication and modeling motivated our decision to create rectangular hinges, and the high prevalence of arc-shaped living hinges in existing compliant mechanisms motivated the \alex{notch} hinges. 


\subsection{Analytical Model from Mechanics Theory}
\label{sec:model}
Nitinol's stress ($\sigma$) as a function of strain ($\epsilon$) is often described using a "bilinear" model due to its phase change from \alex{austenite} to a mixture of \alex{austenite} and martensite at strains above the critical strain: $\epsilon_l$. \alex{Eq. \ref{eq:1} shows this bilinear model.}
\begin{equation}\label{eq:1}
\sigma(\epsilon) = 
    \begin{array}{cc}
  \Bigg\{ & 
    \begin{array}{cc}
      -E\epsilon_l + E_n(\epsilon + \epsilon_l) & \epsilon < \epsilon_l \\
      E\epsilon &  -\epsilon_l \leq \epsilon \leq \epsilon_l \\
      E\epsilon_l + E_n(\epsilon - \epsilon_l) & \epsilon > \epsilon_l
    \end{array}
\end{array}
\end{equation}

\alex{This stress induced transition from austenite to martensite enables the superelastic effect in nitinol above its transition temperature \cite{mwangi_nitinol_2019}.} Above $\epsilon_l$ the elastic modulus of the mixed phase, $E_n$, is lower than the modulus of pure \alex{austenite}, $E$. 
We experimentally quantified the stress-strain curve for our material in Fig. \ref{fig:analysis}f \alex{by performing a tension test on a dogbone sample using a Dynamic Mechanical Analyzer. }

To model the behavior of a rectangular cross section hinge, i.e., compute its bending moment ($M$) as function of hinge angle ($\theta$), we adopt the approach described in York \textit{et al.} 
\cite{york_nitinol_2019}.
\alex{We first compute the strain energy density ($W$) at each point ($x,y$) along the length ($l$) of the hinge assuming pure bending about the neutral axis of a rectangular beam.}
\alex{Next, we integrate over the volume ($V$) of the hinge and use Castigliano's first theorem to obtain moment as a function of hinge angle.}
\begin{equation}\label{eq:5}
    M = \frac{\partial U(\theta)}{\partial \theta} = \frac{\partial}{\partial \theta}\int_{V} W(\epsilon)dV \text{, where }  \epsilon(y,\theta) = \frac{y\theta}{l} 
\end{equation}
\alex{
We can experimentally validate hinge performance by comparing the theoretical predictions (Eq. \ref{eq:5}) with direct measurements of the motor torque required to bend a hinge through the desired angular deflection range using a force sensor (Fig. \ref{fig:analysis}g).}

\subsection{Computational Model in Abaqus}

While the analytical model above works well for rectangular cross sections, it is difficult to adapt for hinges with arbitrary cross sections. 
\alex{Therefore, we implement a nonlinear finite element model in the numerical analysis software Abaqus. We pose the model in 2D under the assumption of planar strain, and exploit symmetry through the center of the hinge to reduce the dimensionality of the model. We use the same bilinear elastic material model as in our theoretical analysis, capturing the desired nonlinear superlastic effects of Nitinol. An encastre boundary condition constrains all degrees of freedom at one end of the hinge. A torque is applied to the opposite end, and the angular displacement at this end is monitored as the torque increases. The analysis considers geometric nonlinearity, and uses a full Newton-Raphson solver to generate solutions with incrementally higher applied torques. Hinge geometry is meshed using 8-node biquadratic quadrilateral plane stress elements with reduced integration (Abaqus type CPS8R), with typical analyses containing 150 elements and 450 degrees of freedom. Mesh is generated automatically using an advancing front algorithm, with mesh density increased near the hinge vertex to accurately resolve stress gradients. Fig. \ref{fig:analysis}d shows the mesh along with the boundary conditions and applied torque. Typical analyses complete in under 10s on a small laptop computer (Intel Core i7-7500U CPU @ 2.70GHz, 16GB RAM). Fig. \ref{fig:analysis}e shows a completed analysis along with the distribution of von Mises stress throughout the hinge.}

 Fig. \ref{fig:analysis}h shows the results from this model compared to physical torque tests for both types of hinge. This model will be incredibly useful for designing hinges with more complex cross sections, which is essential for high-performance devices like the wing mechanism discussed in Section \ref{sec:app}.


\subsection{Experimental Characterization}
\begin{figure*}[htb]
    \centering
    \includegraphics[width=0.9\textwidth]{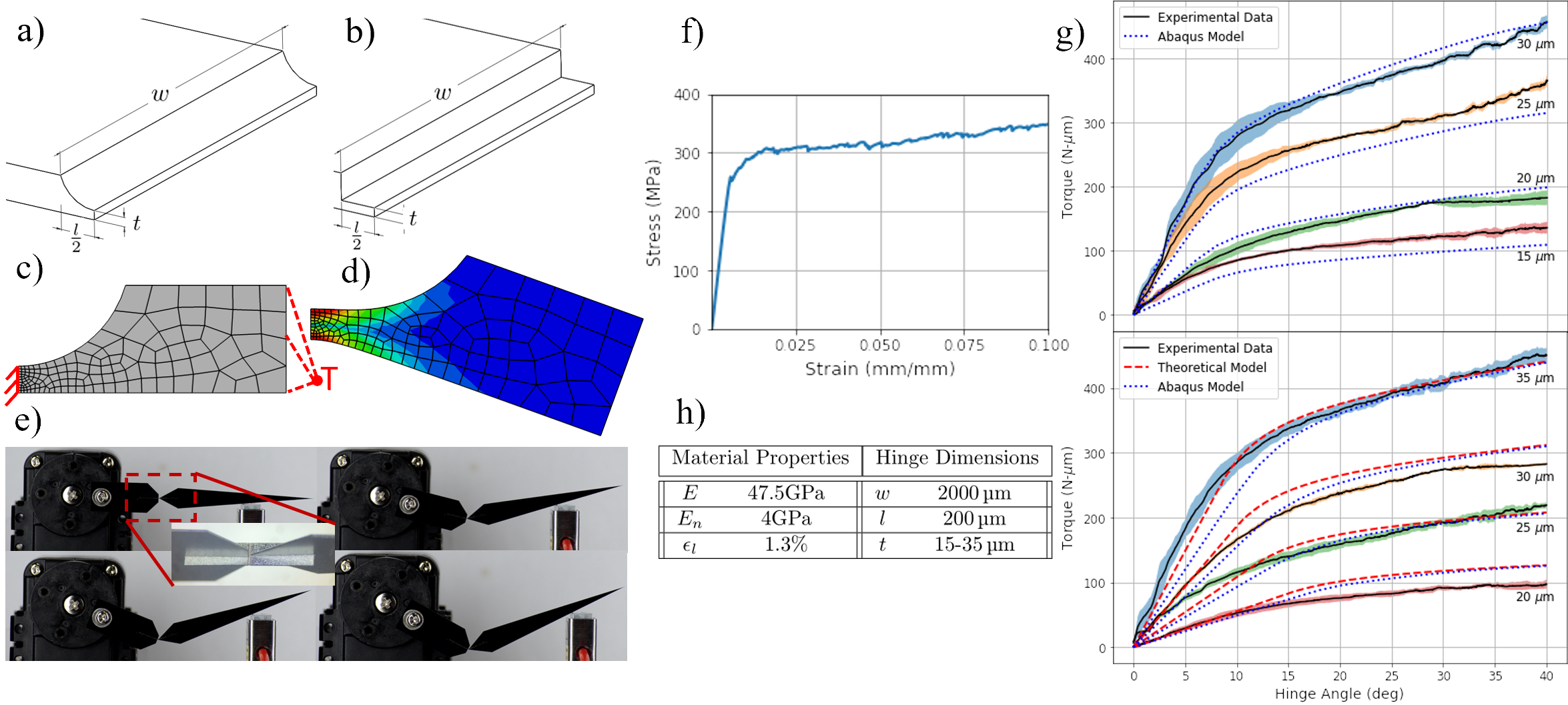}
    \caption{(a) and (b): Diagram of the \alex{notch} and rectangular hinges. Half cross sections are depicted for ease of visualization. (c): Meshed hinge model with boundary conditions and applied torque shown in red. (d): Result of an Abaqus analysis showing the displaced hinge and von Mises stress. (e): Series of 4 images showing the progression of our test for measuring hinge torque. Inset shows a close-up of the hinge from the torque test. We rotated the servo at a speed of 0.05 rad/s to 40 degrees for 8 different hinges. (f): Stress vs. strain data from the nitinol used in this paper. Note the bilinear behavior.  (g): Torque (N-$\mu m$) as a function of hinge angle (deg) for \alex{notch} (top) and rectangular (bottom) hinges with varying hinge thickness. We provide the mean and standard deviation for each of the 8 hinges. (h): Table showing the experimentally determined material properties and dimensions of the hinges.}
    \label{fig:analysis}
\end{figure*}

In order to experimentally validate the theoretical and finite element models for our hinges, we performed a quasi-static test measuring the flexure's torque as a function of angle. The experimental setup, shown in Fig. \ref{fig:analysis}g, includes a servo motor (Dynamixel AX-12A) and a S-Beam type load cell (Futek LSB200) with suitable range for measuring our hinges' torque. We performed 5 trials each for 4 rectangular hinges with thicknesses 35, 30, 25, 20 $\mu m$ and 4 \alex{notch} hinges with thicknesses 30, 25, 20, and 15 $\mu m$, bending each hinge to a 40\textdegree$ $ angle. We passed the data through a 101 order median filter to smooth the noise and took the average and standard deviation of the 5 trials for each thickness.


Fig. \ref{fig:analysis}h displays the results from the torque experiment for the \alex{notch} hinges (left) and the rectangular hinges (right). These experimentally obtained torque profiles show a good agreement with both the theoretical model and the finite element model predictions. We suspect the differences between the experiment and theory are due to minor manufacturing discrepancies, and we discuss ways to reduce these discrepancies in Section \ref{sec:disc}. Having validated the models successfully, we use them for designing living hinges with specifically tuned torque profiles as dictated by the application needs. We explore one such application, a robotic wing, in the next section.


\section{Applications}
\label{sec:app}

We chose to make a 4-bar wing transmission mechanism to illustrate the usefulness of strong, high-displacement hinges for certain milli-robot applications. This design is inspired from the RoboBee wing transmission \cite{wood_design_2007, salem2022flies}. However, our design has about half as many layers because nitinol can be used as a structural material as well as a flexural material. 
The typical operating wing pitch angle for the passive hinge in the mechanism is 50\textdegree  \cite{gravish_anomalous_2016}. \alex{Using equation \ref{eq:5} and an elastic strain limit of 0.8\% for Kapton, we determine that the elastic range for a 200 $\mu m$ long hinge made from 25 $\mu m$ thick Kapton is only about 7.3\textdegree. Thus, existing wing mechanisms made using Kapton as the flexure material operate beyond} the elastic range of the hinges. However, this angle is within the elastic range for nitinol hinges with a thickness of 22 $\mu m$ or less, assuming a elastic strain limit of 6\%. Furthermore, the original mechanism has 15 total layers and 2 different adhesive curing processes. For these reasons, this wing mechanism is ideal for testing our nitinol hinges because we have the potential to greatly increase the mechanism's lifespan while reducing manufacturing time. Using nitinol as both a flexural and structural material in the wing allowed us to reduce the stack to only 7 layers (2 carbon fiber, 2 nitinol, 3 adhesive) and 1 adhesive curing process. 

\begin{figure}
    \centering
    \includegraphics[width=0.45\textwidth]{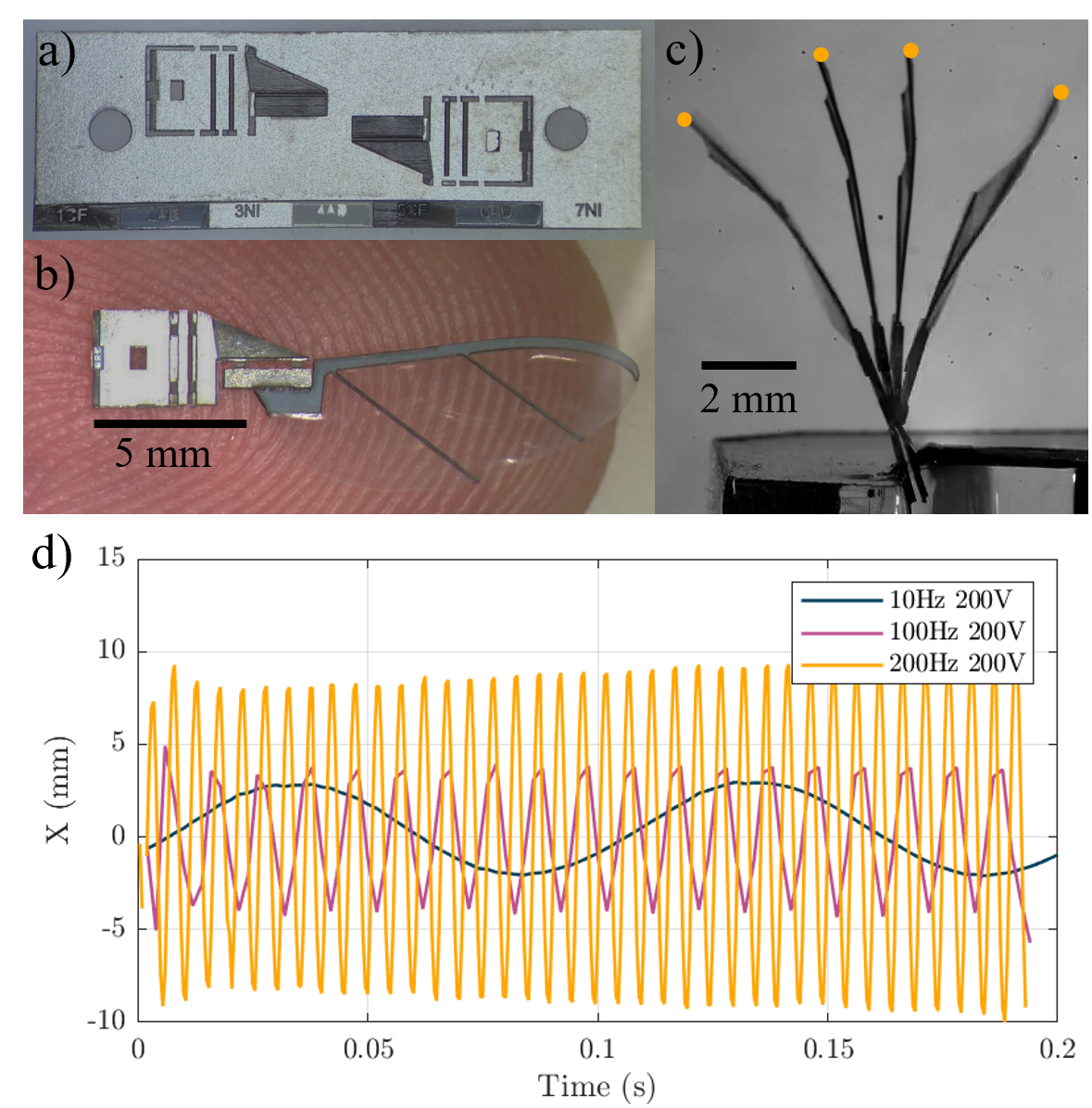}
    \caption{(a): Manufacturing stack-up for the wing mechanism showing the 7 layers. (b) Manufactured wing mechanism on fingertip. (c): Overlay of 4 moments during the wing mechanism test, with t = 0s on the left and t = 2.5ms on the right. The wing had an angle range of 50\textdegree, and the yellow points show the location on the wing that we tracked in the graph below. \alex{(d): Graph of wing tip deflection in the X direction as a function of time for 200V actuator input at 10Hz, 100Hz, and 200Hz.}}
    \label{fig:mechs}
\end{figure}

Fig. \ref{fig:mechs}a shows a stackup of the mechanism's 7-layer design. We cut these 7 layers separately in the laser and aligned them using vertical alignment pins. Next, we heat-pressed them all together in a single cycle. Finally, we performed a cut on the laser to release the mechanism from its supports. The entire manufacturing process takes about three and a half hours. Fig. \ref{fig:mechs}b shows the manufactured wing mechanism resting on a fingertip. We built a small test fixture to hold the wing transmission in place while we actuated it with a piezo-electric actuator. Fig. \ref{fig:mechs}c shows a composite image of the wing during a test in which we drove the piezo-electric actuator at 200 Hz and 200 V. The wing mechanism reached a peak-to-peak stroke amplitude of 50\textdegree, surpassing the elastic range of Kapton. 

However, since this value is within the elastic range for nitinol, we hypothesize that after optimizing the transmissions, they would have a much longer lifespan and demonstrate superior mechanical performance than transmissions made with Kapton as the flexural material. \alex{We also tested the wing at 10 Hz and 100 Hz in addition to the 200 Hz test and graphed the displacement of the wing in Fig. \ref{fig:mechs}d. We used the DLTdv MATLAB app to track the displacement of the wing tip in time \cite{hedrick_software_2008}. The plot in Fig. \ref{fig:mechs}d shows that in order to achieve high displacement, frequency values of 200 Hz or even higher must be used. }

\alex{For future robotic applications of nitinol hinges  designers must acknowledge loading hysteresis and slight residual strains. In high-precision robotic devices, these can be compensated for on the controller side using closed loop control. Furthermore, for legged robots and wing applications, the loading direction matters much more compared to the unloading direction, and the difference between loading and unloading can be effectively ignored.}

\section{Discussion and Future Work}
\label{sec:disc}

Accurately manufacturing 2.5D hinge cross sections with nitinol, while retaining superelastic properties, is essential for nitinol-based applications in micro-robots and medical devices. We have demonstrated that this is possible using femtosecond laser technology. \alex{It is important to note that while the femtosecond laser method minimizes HAZ, we still noticed slight debris and recasting from the ablated material \cite{mwangi_nitinol_2019}. This recast material does not affect the transient performance of the hinge, as illustrated by the accuracy of the material models in Fig. \ref{fig:analysis}. Instead, the recast layer degrades the surface finish of the hinges thus reducing the hinge lifetime. \alexnew{We performed preliminary fatigue experiments on the hinges and found that a 25 $\mu m$ thick notch hinge broke at around 2000 cycles when loaded and unloaded to 36$^{\circ}$ which is on par with previous studies \cite{york_nitinol_2019}}. However, solving the problem of recast material requires additional techniques like polishing or machining underwater. We will explore laser polishing techniques to extend the lifetime of nitinol hinges as immediate next steps \cite{mwangi_nitinol_2019}. }
These improvements will be vital for future applications which we will actively actuate and control using piezoelectrics or other milli-device compatible actuation methods. Furthermore, due to the accuracy of the manufacturing process, we will attempt to prototype monolithic 2-DOF flexures. Multi-DOF flexures have the potential to further reduce size, weight, and complexity of miniature devices.

\alex{As discussed in the Introduction, nitinol hinges also have the potential to minimize the negative effects of off-axis loading because our manufacturing method supports 3D contouring of hinges. Off-axis compliance is a problem for existing microrobotic hinges which are typically made from polyimide \cite{wood_microrobot_2008}. Some solutions to limiting off-axis compliance include castellated hinges \cite{doshi_model_2015} and "no-buckling" flexure designs that always have part of the flexure in tension \cite{wood_microrobot_2008}. However, castellation stiffens the hinge which may be unwanted for some applications and no-buckling hinges reduce the hinge's effective width. Future nitinol hinges manufactured with a femtosecond laser could have varying cross-section across the width thus reducing the significance of off-axis compliance.}

Finally, we believe we can reduce the tolerance for 2.5D hinge cuts below $\pm 5 \mu m$ by varying the laser's attenuation settings. By dialing in exact power settings using the confocal microscope to get even more accurate cut settings, we can slice the hinge with smaller step sizes, allowing for tighter tolerances. 

In conclusion, we proposed a new process for manufacturing nitinol living hinges: femtosecond laser machining. We characterized this process in order to accurately manufacture complex 2.5D hinge shapes. We then tested the hinge torque, compared the results to a theoretical model, and built a prototype device. With future supplementary research into optimal hinge cross sections, these versatile hinges will be useful in a wide number of applications across multiple fields of mechanical engineering including microrobotics.

\section*{Acknowledgments}

The authors would like to thank William McDonnell for his assistance with the various test setups, Brandon Hayes for help obtaining the confocal microscope images, COSINC staff members Dr. Nicholas Weadock, Dr. Adrian Gestos, and Dr. Tomoko Borsa for advice and for obtaining the XRD results and Dr. Stephen Uhlhorn (6D Lasers Inc.) for femtosecond photomachining advice.

\bibliographystyle{ieeetran}
\bibliography{references_v1}

\end{document}